# ARTIST, STYLE AND YEAR CLASSIFICATION USING FACE RECOGNITION AND CLUSTERING WITH CONVOLUTIONAL NEURAL NETWORKS

Doruk Pancaroglu

STM A.S., Ankara, Turkey

*ABSTRACT*

*Artist, year and style classification of fine-art paintings are generally achieved using standard image classification methods, image segmentation, or more recently, convolutional neural networks (CNNs). This works aims to use newly developed face recognition methods such as FaceNet that use CNNs to cluster fine-art paintings using the extracted faces in the paintings, which are found abundantly. A dataset consisting of over 80,000 paintings from over 1000 artists is chosen, and three separate face recognition and clustering tasks are performed. The produced clusters are analyzed by the file names of the paintings and the clusters are named by their majority artist, year range, and style. The clusters are further analyzed and their performance metrics are calculated. The study shows promising results as the artist, year, and styles are clustered with an accuracy of 58.8, 63.7, and 81.3 percent, while the clusters have an average purity of 63.1, 72.4, and 85.9 percent.*

*KEYWORDS*

*Face Recognition, Clustering, Convolutional Neural Networks, Art Identification*

## 1. INTRODUCTION

Art classification, or more correctly, painting classification in the context of this paper, is a problem concerned about correctly identifying a piece of art's creator, the artistic movement it belongs to, and its approximate age. Pieces of art stored in museums are identified and categorized manually by art experts and curators. As in all things that involve humans, this is a very error-prone process. There are a lot of cases of art fraud involving museums, auctions, and large sums of money being paid for worthless reproductions [1].

Art classification by computers is an active area of research because paintings of similar subjects (still lives, for example) by different painters can have very different styles, leading to varied classification results. A solution to the problem of art classification can also find use in online museums, educational purposes, and recommendation systems. Currently, most solutions to the problem of art classification are based on image segmentation [2] or stochastic modeling [3]. Image segmentation is especially useful in classifying modern or abstract art. However, this paper aims to use the methods developed for face recognition to solve the problem of art classification.

Many paintings, excluding the ones with the abstract and modern styles (which will not be featuring in this work obviously), include faces. A large percentage of paintings are portraits, as they were the bread and butter of painters [4]. A selection of these works can be observed in Figure 1.

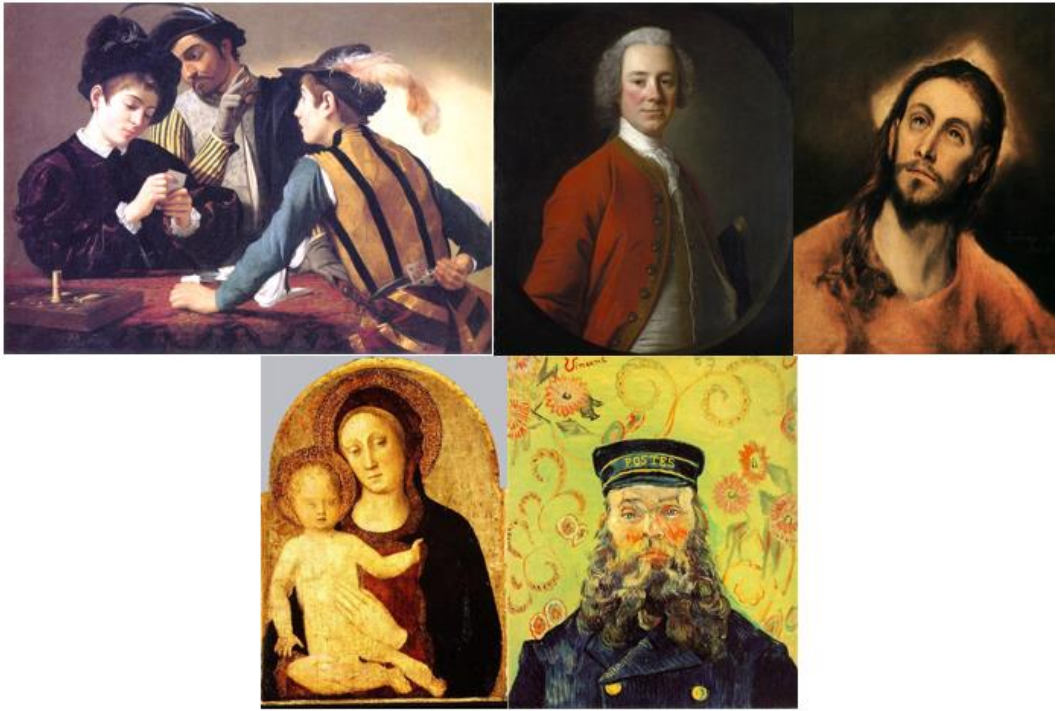

Figure 1: Five paintings from different artists, eras and artistic styles. Note the difference in the styles of faces, which would prove beneficial in clustering artists, years and styles.

The abundance of faces in the paintings and the fact that many painters have a distinct style of painting can enable face detection and clustering methods to detect artists from the faces in their paintings. Moreover, the style of faces can also be used to classify the era in which the painting belongs. The field of image processing has attained great momentum with the introduction of convolutional neural networks (CNNs). Major social media and technology companies like Google and Facebook have invested in face detection, recognition, and clustering methods that use CNNs, such as DeepFace [5] and FaceNet [6].

The aim of this paper is to overcome the problem of art classification with face detection and clustering methods that are using convolutional neural networks (CNNs). The dataset that will be used in this work is named WikiArt Dataset [7], which contains paintings gathered from WikiArt [8], a website with a large number of labeled art objects from many different artists and eras.

This paper is organized into six sections. In the first section, the problem is introduced and some background information is given. In the second section, related work about the problem will be discussed in detail. In the third section, the dataset used in this work will be presented. In the fourth section, the implementation of the work will be explained. In the fifth section, the results of the work will be presented. Finally, in the sixth section, the work will be concluded and future directions will be discussed.

## 2. RELATED WORK

Using computers in the classification of art objects is a relatively new field, but some of the more important achievements will be explained in this section.

Before using CNNs for image processing became popular, two works, the first by S. Lyu, D. Rockmore and H. Farid in 2004 [9], and the second by C. Johnson et al. in 2008 [10], aimed to identify and authenticate art by analyzing the brush strokes of the paintings. Many famous artists have distinct brush strokes which are very important to ascertain if the painting in question is a fake or not. These works use wavelet analysis of the brushstrokes to find out if a painting is an original or a reproduction.

Another work, created by C. Li and T. Chen in 2009 [11] propose a method to evaluate the aesthetic quality of a given piece of art. First, the training set is rated by 23 humans to create a baseline for aesthetic quality. Then, naïve Bayesian and adaptive boosting (Adaboost [12]) classifiers are used to classify the test set. Finally, the classified images are compared to the human scores and whether a painting is found aesthetic or not is found out.

More recent works that use CNN-backed methods have also appeared. A paper by S. Karayev et al., published in 2013 [13], used CNN-based stochastic gradient descent classifiers to identify the artistic style of a set of paintings. The results are compared with style labels given to the painting by humans.

In 2014, authors T. Mensink and J. V. Gemert published a paper [14] consisting of four challenges accompanied by a dataset of over 110,000 images of pieces of art located in the Rijksmuseum in Amsterdam, Netherlands. The four challenges were predicting the artist, predicting the type of the art (painting, sculpture, etc.), predicting the material, and lastly, predicting the year of creation.

The Rijksmuseum Challenge paper proposes baseline experiments for the four challenges as well. These experiments use 1-vs-Rest linear SVM classifiers. The dataset itself has images encoded with Fisher Vectors [15] that are aggregating local SIFT descriptors, embedded in a global feature vector.

Another work by L. A. Gatys, A. S. Ecker, and M. Bethge, published in 2015 [16], uses VGG-Network, a CNN based classifier to create a method that fuses a given photograph with a painting and creates art. This opens up a different field altogether: can a machine create art?

In 2016, W. R. Tan, C. S. Chan, H. E. Aguirre, and K. Tanaka published a paper [17], in which paintings are classified using CNNs with respect to their style, genre, and artists. The work had two objectives. Firstly, the work aimed to train a CNN model as a proof-of-concept for art classification. Secondly, the work aimed to be able to classify modern and abstract art, and tried to find an answer to the question: "is a machine able to capture imagination?"

One of the motivations for this paper stems from Google Arts & Culture [18]. Created in 2011, Google Arts & Culture is an online platform functioning as a museum, where partner museums of Google contribute their collections for online touring. In 2018, an extension to the mobile application of Google Arts & Culture appeared, in which the user's selfie would be matched with a portrait stored in the databases of Google Arts & Culture. For this, Google uses its own CNN face recognition method, FaceNet, which will be explained in detail in the following paragraph.

FaceNet is a face recognition method developed by researchers from Google, F. Schroff, D. Kalenichenko, and J. Philbin in 2015 [6]. FaceNet uses deep convolutional networks that are trained for direct optimization for embedding, which itself is the process of measuring the facial similarities between two images.

FaceNet also bypasses the bottleneck layer found in other CNN face recognition methods and instead, it trains the output as a compact 128-D embedding using a triplet-based loss function. FaceNet is also touted as a "pose-invariant" face recognizer, which is a big advantage for classifying paintings as well.

FaceNet trains CNNs with Stochastic Gradient Descent [20] with standard backpropagation and AdaGrad [21]. Two types of architectures are proposed for FaceNet.

The first one is a Zeiler&Fergus [22] architecture with a model consisting of 22 layers. This architecture has 140 million parameters and it needs a computing power of 1.6 billion FLOPS for each image.

Table 1: The structure of the Zeiler&Fergus architectural model for FaceNet, with 1x1 convolutions. The input and output columns are represented as row x col x #filters. The kernel column is represented as row x col x stride.

| Layer  | Input       | Output      | Kernel      | Params. | FLOPS |
|--------|-------------|-------------|-------------|---------|-------|
| conv1  | 220x220x3   | 110x110x64  | 7x7x3,2     | 9K      | 115M  |
| pool1  | 110x110x64  | 55×55×64    | 3x3x64,2    | 0       |       |
| rnorm1 | 55×55×64    | 55×55×64    |             | 0       |       |
| conv2a | 55×55×64    | 55×55×64    | 1x1x64,1    | 4K      | 13M   |
| conv   | 55×55×64    | 55×55×192   | 3x3x64,1    | 111K    | 335M  |
| rnorm2 | 55×55×192   | 55×55×192   |             | 0       |       |
| pool2  | 55×55×192   | 28×28×192   | 3x3x192,2   | 0       |       |
| conv3a | 28×28×192   | 28×28×192   | 1x1x192,1   | 37K     | 29M   |
| conv3  | 28×28×192   | 28×28×384   | 3x3x192,1   | 664K    | 521M  |
| pool3  | 28×28×384   | 14×14×384   | 3x3x384,2   | 0       |       |
| conv4a | 14×14×384   | 14×14×384   | 1x1x394,1   | 148K    | 29M   |
| conv4  | 14×14×384   | 14×14×256   | 3x3x384,1   | 885K    | 173M  |
| conv5a | 14×14×256   | 14×14×256   | 1x1x256,1   | 66K     | 13M   |
| conv5  | 14×14×256   | 14×14×256   | 3x3x256,1   | 590K    | 116M  |
| conv6a | 14×14×256   | 14×14×256   | 1x1x256,1   | 66K     | 13M   |
| conv6  | 14×14×256   | 14×14×256   | 3x3x256,1   | 590K    | 116M  |
| pool4  | 14×14×256   | 7×7×256     | 3x3x256,2   | 0       |       |
| concat | 7×7×256     | 7×7×256     |             | 0       |       |
| fc1    | 7×7×256     | 1×32×128    | Maxout p=2  | 103M    | 103M  |
| fc2    | 1×32×128    | 1×32×128    | Maxout p=2  | 34M     | 34M   |
| fc7128 | 1×32×128    | 1×1×128     |             | 524K    | 0.5M  |
| l2     | 1×1×128     | 1×1×128     |             | 0       |       |
| Total  |             |             |             | 140M    | 1.6B  |

Computer Science & Information Technology (CS & IT)                                           45Table 2: The structure of the inception architectural model for FaceNet

| Type | Output Size | Depth | #1x1 | #3x3 reduce | #3x3 | #5x5 reduce | #5x5 | Pool proj (p) | Params. | FLOPS |
|---|---|---|---|---|---|---|---|---|---|---|
| conv1 (7x7x3,2) | 112x112x64 | 1 | | | | | | | 9K | 119M |
| max pool + norm | 56x56x64 | 0 | | | | | | m 3x3,2 | | |
| inception (2) | 56x56x192 | 2 | | 64 | 192 | | | | 115K | 360M |
| norm + max pool | 28x28x192 | 0 | | | | | | m 3x3,2 | | |
| inception (3a) | 28x28x256 | 2 | 64 | 96 | 128 | 16 | 32 | m, 32p | 164K | 128M |
| inception (3b) | 28x28x320 | 2 | 64 | 96 | 128 | 32 | 64 | $L_2$, 64p | 228K | 179M |
| inception (3c) | 14x14x640 | 2 | 0 | 128 | 256,2 | 32 | 64,2 | m 3x3,2 | 398K | 108M |
| inception (4a) | 14x14x640 | 2 | 256 | 96 | 192 | 32 | 64 | $L_2$, 128p | 545K | 107M |
| inception (4b) | 14x14x640 | 2 | 224 | 112 | 224 | 32 | 64 | $L_2$, 128p | 595K | 117M |
| inception (4c) | 14x14x640 | 2 | 192 | 128 | 256 | 32 | 64 | $L_2$, 128p | 654K | 128M |
| inception (4d) | 14x14x640 | 2 | 160 | 144 | 288 | 32 | 64 | $L_2$, 128p | 722K | 142M |
| inception (4e) | 7x7x1024 | 2 | 0 | 160 | 256,2 | 64 | 128,2 | m 3x3,2 | 717K | 56M |
| inception (5a) | 7x7x1024 | 2 | 384 | 192 | 384 | 48 | 128 | $L_2$, 128p | 1.6M | 78M |
| inception (5b) | 7x7x1024 | 2 | 384 | 192 | 384 | 48 | 128 | m, 128p | 1.6M | 78M |
| avg pool | 1x1x1024 | 0 | | | | | | | | |
| fully conn | 1x1x128 | 1 | | | | | | | 131K | 0.1M |
| $L_2$ normalization | 1x1x128 | 0 | | | | | | | | |
| Total | | | | | | | | | 7.5M | 1.6B |

The second architecture is based on the GoogleNet style Inception Models [23]. This architecture is 17 layers deep. This architecture is composed of 7,5 million parameters and it consumes less computing power compared to the first one.

Detailed information about these two architectures can be found in Tables 1 and 2.

FaceNet achieved high results, even higher than humans in well-known facial recognition benchmarks such as Labelled Faces in the Wild (LFW) [24] and Youtube Faces Database (YDB) [25]. These results can be seen in Table 3.

Table 3: The accuracy values of FaceNet and other prominent face recognition methods tested with the datasets LFW and YDB. The values denoted in bold are the highest scores.

| Method | Accuracy (LFW) | Accuracy (YDB) |
|---|---|---|
| FaceNet | **0.9963** | **0.9512** |
| Humans | 0.9920 | - |
| DeepFace | 0.9735 | 0.9140 |
| Joint Bayesian [26] | 0.9633 | - |
| DDML[27] | 0.9068 | 0.8230 |
| LM3L [28] | 0.8957 | 0.8130 |
| Eigen-PEP [29] | 0.8897 | 0.8480 |
| CNN-3DMM [30] | 0.8880 | 0.9235 |
| APEM (fusion) [31] | 0.8408 | 0.7910 |

There are several code implementations of FaceNet available on GitHub. The one created by David Sandberg [35] is selected for its ease of use and better documentation.

## 3. DATASET

The dataset used in this work is named WikiArt Dataset. The dataset is created for the paper published by B. Saleh and A. Elgammal in 2015 [32]. The work itself is an SVM-based art classification method.

The dataset is created with the paintings collected from the WikiArt website. WikiArt is arguably the largest online and free collection of digitized paintings.

The dataset contains 81,479 paintings from 2,148 artists. The paintings are also categorized into styles from different periods of art history, totaling 27. Lastly, paintings are categorized into 45 genres such as still lives or portraits.

In the work by B. Saleh and A. Elgammal, The dataset is grouped into three different subsets for three distinct classifications challenges: style identification, genre identification, and artist identification.

For the style identification challenge, the dataset is subdivided into 27 styles with at least 1,500 paintings each, with a total of 78,449 paintings. For the genre identification challenge, the dataset is subdivided into 10 genres with at least 1,500 paintings each, totaling 63,691 paintings.

For the artist identification challenge, the dataset is subdivided into 23 artists with at least 500 paintings each, with a total of 18,599 paintings.

The detailed styles, genres, and artists (25 of the 1,119) groupings of the WikiArt dataset can be seen in Table 4.

For this paper, the styles that do not have any faces (action paintings, abstract expressionism) or a majority of faces in their paintings (color field painting, pop-art), or with faces that are not suitable for recognition (cubism) are removed from the dataset. This reduces the total number of paintings in the dataset to 67,064, with 16 styles and 1,382 artists.

Table 4: List of styles, genres (not used in the scope of this work) and an incomplete list of painters present in the WikiArt dataset. The styles marked with (*) are omitted from this work because of the lack of faces or the faces being distorted.

| Classification Task | List of Members |
|---|---|
| Style | Abstract Expressionism(*); Action Painting(*); Analytical Cubism(*); Art Nouveau Modern Art; Baroque; Color Field Painting(*); Contemporary Realism; Cubism(*); Early Renaissance; Expressionism; Fauvism(*); High Renaissance; Impressionism; Mannerism-Late-Renaissance; Minimalism(*); Primitivism-Naive Art(*); New Realism; Northern Renaissance; Pointillism; Pop Art(*); Post Impressionism; Realism; Rococo; Romanticism; Symbolism; Synthetic Cubism(*); Ukiyo-e(*) |
| Genre | Abstract painting; Cityscape; Genre painting; Illustration; Landscape; Nude painting; Portrait; Religious painting; Sketch and Study; Still Life |
| Artist (Selected) | Leonardo Da Vinci, Michelangelo, Caravaggio, Diego Velazquez, El Greco; Albrecht Durer; Francisco Goya; Boris Kustodiev; Camille Pissarro; Claude Monet; Edgar Degas; Eugene Boudin; Gustave Dore; Ilya Repin; Ivan Aivazovsky; Ivan Shishkin; John Singer Sargent; Marc Chagall; Nicholas Roerich; Pablo Picasso; Paul Cezanne; Pierre-Auguste Renoir; Rembrandt; Salvador Dali; Vincent van Gogh |

## 4. IMPLEMENTATION

As mentioned in section 2, David Sandberg's FaceNet implementation is selected to run FaceNet's face clustering method with WikiArt dataset. This implementation works with Tensorflow [33], an open-source machine learning framework with CUDA support. The implementation is written using Python.

For computational purposes, paintings from the following styles are selected with all their available artists: Early Renaissance, High Renaissance, Late Renaissance, Northern Renaissance, Baroque and Rococo. These 6 styles amount to 12,907 paintings and 192 artists.

### 4.1. Pre-processing

Firstly the preprocessing phase is done. This phase is called the alignment phase in the implementation, meaning that only the face and a given margin is extracted from a larger image. A sample process can be seen in Figure 2.

In our case, the face or if present, multiple faces are extracted from the paintings. Extracting multiple faces from a painting provides better insight into an artist's style of painting a face.
In the alignment phase, the output image size is selected as 160x160 pixels and the margin for the area around the bounding box of the face is 32 pixels.

### 4.2. Training

Secondly, the training phase begins. For this phase, the training model is obtained by training the classifier on the VGGFace2 dataset [34] created by Q. Cao, L. Shen W. Xie, O. M. Parkhi, and A. Zisserman. This dataset contains 3.3 million images for over 9,000 people.

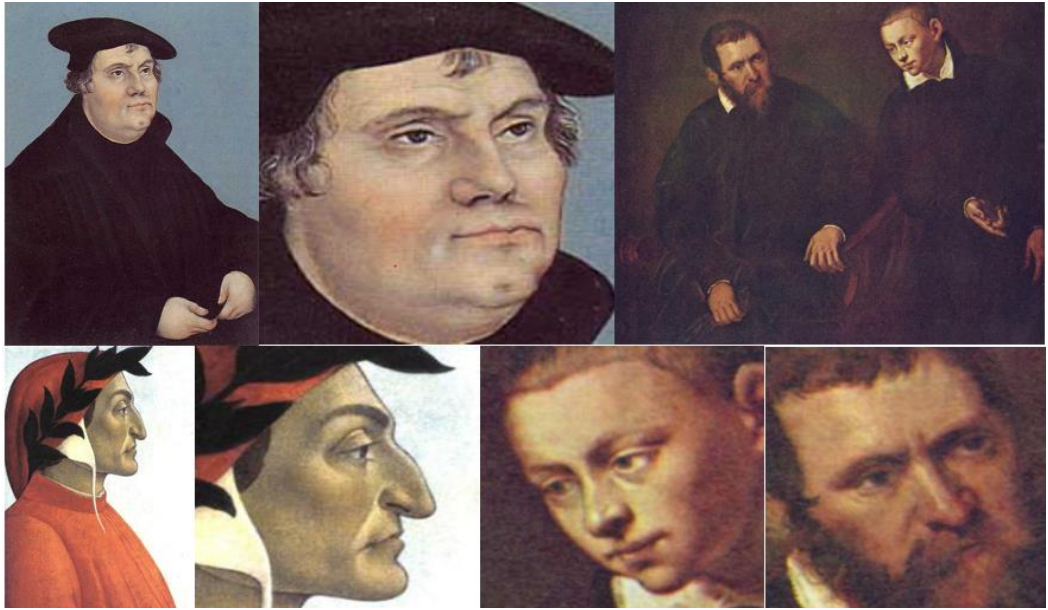

Figure 2: The results of the face extraction process used in three different paintings. The extraction process also works for multiple faces in the paintings.

### 4.3. Clustering

For the clustering operation itself, DBSCAN [34] algorithm, created by J. Sander, M. Ester, H. P. Kriegel, and X. Xu, is used. The minimum number of required images to form a cluster set at 25. The clustering operation runs the FaceNet face recognition method and clusters the similar faces, based on the Euclidean distance matrix.

### 4.3. Analyzing

Following the clustering operation, the produced clusters are analyzed using the file names of the paintings in the clusters. The file names contain the artist's name, the name of the painting, the style of the painting, and the year of completion (it should be noted that in the dataset, not all paintings have a year of completion).

After the file names are parsed, the second part of the analyzing phase starts. In this phase, the clusters are named according to the dominant artist, style, or year. If no dominant artist, style, or year is found in a cluster, that cluster is omitted from the results.

To achieve a better image classification accuracy in terms of years, the years of the paintings are grouped in 50-year periods (e.g., if a cluster has a majority of paintings dated from years 1500 to 1550, that cluster is named 1500-1550).

### 5. RESULTS AND EVALUATION

In the evaluation of this work, all the clusters are evaluated three times for the three tasks of classification: artist classification, style classification, and year classification.

Four different values make up the formulas of the results: True Positive (TP), False Positive (FP), True Negative (TN), and False Negative (FN).

In the context of this work and the produced clusters, TP is the total number of paintings that are present in their correct clusters (i.e. the paintings belonging to the majority artist of a cluster). FP is the total number of paintings that are present in the cluster, but not belonging to the majority artist of that cluster. TN is the total number of paintings that are not present in the cluster, and also not belonging to the majority artist of the cluster. FN is the total number of paintings that are not present in the cluster, but belonging to the majority artist of that cluster. The results are separated into two groups, cluster-specific and inter-cluster.

## 5.1. Cluster-Specific Results

Four different metrics are calculated for each cluster. The metrics in question are accuracy, precision, recall, and f-measure. The accuracy metric is the general rate of the correctness of the cluster's artist compared to the whole dataset. As this metric does not give a meaningful explanation of the results by itself, other metrics are also used.

The metric of precision is the rate of the correctness the cluster's artist in a given cluster, while recall is the rate of the correctness of that cluster's artist in the whole dataset. F-measure, or F1 score, is a measure that combines precision and recall, the harmonic mean of the two.
The formulas can be observed in Equations (1), (2), (3), and (4).

$$Accuracy = \frac{TP+TN}{TP+TN+FP+FN} \quad (1)$$

$$Precision = \frac{TP}{TP+FN} \quad (2)$$

$$Recall = \frac{TP}{TP+FP} \quad (3)$$

$$F - Measure = 2 \times \frac{Precision \times Recall}{Precision + Recall} \quad (4)$$

A selection of the results of the formed clusters which have a majority, separated by their tasks can be observed in Tables 5, 6, and 7. Considering the large number of artists, styles, and year periods in the dataset, not all clusters produced in the classification tasks are present in these tables.

## 5.2. Inter-Cluster Results

In addition to the cluster-specific results, metrics concerning all the created clusters are calculated. These metrics are purity, normalized mutual information (NMI), and Rand Index (RI).

Purity is a straightforward measure, similar to accuracy that produces a general quality of clustering. But for clusters with single items, purity tends to produce misleading results. The problem is solved by using normalized mutual information. NMI is useful at producing normalized values for evaluating the quality of clustering.

The final metric, Rand Index is a measure of similarity between clusters. RI is calculated by adding the pairs of total TP and TN values in every cluster and dividing this value by the total number of images.

These formulas can be observed in Equations (5), (6), and (7).

$$Purity(\Omega, \Phi) = \frac{1}{N} \sum_k max_j |\omega_k \cap \varphi_j| \quad (5)$$

$$NMI(\Omega, \Phi) = \frac{I(\Omega; \Phi)}{|H(\Phi) + H(\Phi)|/2} \quad (6)$$

$$RI = \frac{\sum_1^\Omega TP + TN}{TP + TN + FP + FN} \quad (7)$$

The inter-cluster results, separated by their tasks can be observed in Table 8.

Table 5: The results of some of the formed artist clusters with a majority

| Artist Cluster | Accuracy | Precision | Recall | F-Measure |
|---|---|---|---|---|
| Rembrandt | 0.759 | 0.949 | 0.691 | 0.800 |
| Durer | 0.672 | 0.932 | 0.614 | 0.740 |
| Rubens | 0.620 | 0.749 | 0.456 | 0.567 |
| J. Reynolds | 0.620 | 0.979 | 0.312 | 0.473 |
| El Greco | 0.615 | 0.994 | 0.314 | 0.477 |
| Velazquez | 0.585 | 0.903 | 0.274 | 0.420 |
| T. Gainsborough | 0.583 | 0.983 | 0.269 | 0.422 |
| Tintoretto | 0.574 | 0.829 | 0.312 | 0.454 |
| Bosch | 0.549 | 1.000 | 0.285 | 0.444 |
| Raphael | 0.541 | 1.000 | 0.292 | 0.453 |
| Caravaggio | 0.490 | 1.000 | 0.141 | 0.248 |
| Botticelli | 0.453 | 0.853 | 0.185 | 0.304 |
| Average | 0.588 | 0.931 | 0.345 | 0.483 |

Table 6: The results of some of the formed style clusters with a majority

| Style Cluster | Accuracy | Precision | Recall | F-Measure |
|---|---|---|---|---|
| Baroque | 0.759 | 0.741 | 0.833 | 0.784 |
| Baroque | 0.708 | 0.955 | 0.587 | 0.727 |
| Baroque | 0.694 | 0.909 | 0.590 | 0.715 |
| Baroque | 0.672 | 0.919 | 0.549 | 0.687 |
| Rococo | 0.657 | 0.994 | 0.452 | 0.622 |
| Northern Ren. | 0.647 | 0.902 | 0.442 | 0.593 |
| Baroque | 0.622 | 0.719 | 0.537 | 0.615 |
| Northern Ren. | 0.618 | 0.996 | 0.426 | 0.596 |
| Northern Ren. | 0.616 | 0.957 | 0.426 | 0.590 |
| Mannerism | 0.574 | 0.988 | 0.279 | 0.435 |
| Rococo | 0.567 | 0.739 | 0.347 | 0.473 |
| Rococo | 0.511 | 0.574 | 0.259 | 0.357 |
| Average | 0.637 | 0.866 | 0.477 | 0.599 |

Table 7: The results of some of the formed clusters with a majority

| Year Cluster | Accuracy | Precision | Recall | F-Measure |
|---|---|---|---|---|
| 1600-1650 | 0.904 | 0.996 | 0.899 | 0.945 |
| 1650-1700 | 0.858 | 0.950 | 0.860 | 0.903 |
| 1600-1650 | 0.856 | 0.964 | 0.867 | 0.913 |
| 1600-1650 | 0.856 | 0.969 | 0.865 | 0.914 |
| 1700-1750 | 0.820 | 0.998 | 0.767 | 0.867 |
| 1700-1750 | 0.816 | 0.989 | 0.757 | 0.857 |
| 1600-1650 | 0.815 | 0.904 | 0.867 | 0.885 |
| 1600-1650 | 0.812 | 0.901 | 0.863 | 0.882 |
| 1650-1700 | 0.809 | 0.978 | 0.792 | 0.875 |
| 1550-1600 | 0.802 | 0.953 | 0.789 | 0.863 |
| 1500-1550 | 0.766 | 0.895 | 0.798 | 0.844 |
| 1500-1550 | 0.649 | 0.756 | 0.754 | 0.755 |
| Average | 0.813 | 0.938 | 0.823 | 0.875 |

Table 8: The results of the cluster groups produced by their tasks. Note that the number of style and year clusters are much higher than the number of distinct styles and year-periods of the dataset. Thus, different clusters are treated as one when calculating accuracy.

| Cluster | # of Clusters | Accuracy | Purity | NMI | RI |
|---|---|---|---|---|---|
| Artist | 115 | 0.598 | 0.631 | 0.394 | 0.489 |
| Style | 86 | 0.666 | 0.724 | 0.407 | 0.563 |
| Year | 30 | 0.875 | 0.859 | 0.440 | 0.645 |
| Average | - | 0.712 | 0.738 | 0.413 | 0.565 |

## 5.3. Evaluation

The results will be explained and evaluated according to the three classification tasks.

The artist classification task produced 126 clusters. 115 of these clusters had a majority artist. These clusters produced an average accuracy, precision, recall, and f-measure values of 58.8%, 93.1%, 34.5%, and 48.3% respectively. The artist clusters, on the whole, are created with the purity, NMI and RI values of 63.1%, 39.4%, and 48.9% respectively.

The results of the artist classification clusters were not strictly high. This can be attributed to the selected styles which include a lot of religious paintings. This caused one of the clusters to have a majority of Jesus Christ faces painted by different artists.

It can also be seen that artists from the early renaissance were harder to cluster while later periods such as baroque and rococo produced clusters with more defined artists and higher metrics. Nevertheless, clusters of artists with a distinct style such as Rembrandt, Dürer, and El Greco have relatively higher scores.

In terms of style, 88 clusters in total were created, with 86 of them having a majority of paintings belonging to the same artistic style. These clusters produced an average accuracy, precision, recall, and f-measure values of 63.7%, 86.6%, 47.7%, and 59.9% respectively. The style clusters, on the whole, are created with the purity, NMI and RI values of 72.4%, 40.7%, and 56.3% respectively.

An interesting case in this is that while only 6 different styles from the dataset were used for clustering, numerous different clusters would have a majority of the same style. This can be attributed to the fact that the dataset is composed of paintings produced roughly between the years 1400 and 1750, and a large part of that year interval has paintings that are more or less similar in style. The true difference in artistic styles started to occur in the 19$^{th}$ century.

In terms of years, 74 clusters in total were created, with 30 of them having a majority. The large difference between the created clusters and the ones with majorities is because the need for a majority of 50-year periods is lacking in most of them. These clusters produced an average accuracy, precision, recall, and f-measure values of 81.3%, 93.8%, 82.3%, and 87.5% respectively. The style clusters, on the whole, are created with the purity, NMI, and RI values of 87.5%, 44.0%, and 64.5% respectively.

Similar, but not as abundant as the style clusters, year clusters also had multiple majorities that pointed to the same 50-year periods. This can be attributed to the difference in parameters used in the artist clustering and the style and year clustering tasks.

It should be noted that using the same parameters (especially higher threshold) for both tasks did not necessarily lead to a lower number of clusters, which would ideally be 6 for styles and 10 for 50-year periods. It can be surmised that for a more accurate year and style classifications, more refined methods are needed.

## 6. CONCLUSION

It can be said that classifying fine-art paintings by extracting the faces in them, and running face detection methods with CNNs is a novel and promising approach.

While artist classification performance is lower than style and year classification performances, this is understandable considering the explanations given in the evaluation part of chapter 5.

Not withholding the fact that this approach works only with paintings that include a face or faces, and thus it is unable to be of any use for the works of many great artists who do not paint faces, this type of approach would still be useful in a variety of situations such as art recommendation and educational purposes.

Future directions of this work include using the whole WikiArt dataset for more refined results, using other prominent art datasets, and implementing different face recognition and clustering methods to compare their performances. Combining this approach with other art classification solutions (analyzing brush strokes, for example) would solve some of the shortcomings of the work, such as the similar art styles of the earlier eras. Finally, creating bigger clusters for style and year classification tasks is another future objective.

Computer Science & Information Technology (CS & IT)                                             53

**AUTHORS**

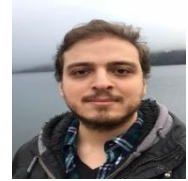

**Doruk Pancaroglu** has been working at STM A.S. as a senior software engineer since 2012. He is a PhD student of computer engineering at Hacettepe University since 2016. He obtained his BSc for Computer Engineering from Sabanci University in 2010 and his MSc for Computer Engineering from TOBB ETU in 2014. Since 2012, he has been working at STM A.S. as a senior software engineer. His research interests are network security, internet of things and machine learning, with a focus on IoT security.